\documentclass[runningheads]{llncs}
\usepackage{graphicx}
\usepackage{amssymb}
\usepackage{enumitem}
\usepackage{mathtools}
\usepackage{array}
\usepackage{verbatim}
\usepackage{pgfgantt}

\usepackage{pgfplots}
\pgfplotsset{width=4.3cm,compat=1.6}
\usepackage{pgfplotstable}
\usepackage[caption=true]{subfig}
\usepackage{multirow}
\usepackage{hyperref}
\usepackage{textpos}

\begin{document}
\title{\textsc{JPLink}: On Linking Jobs to Vocational Interest Types\thanks{This research was supported by the National Research Foundation, Prime Minister’s Office, Singapore under its International Research Centres in Singapore Funding Initiative.}}
\author{Amila Silva\inst{1} \and
Pei-Chi Lo\inst{2} \and
Ee-Peng Lim\inst{2}}
\authorrunning{A. Silva et al.}
\institute{School of Computing and Information Systems, The University of Melbourne\\
\email{amila.silva@student.unimelb.edu.au}\and
School of Information System, Singapore Management University
\email{\{pclo.2017@phids.,eplim@\}smu.edu.sg}}
\begin{textblock*}{100mm}(0.05\textwidth,-3.5cm)
\centering \large Accepted as a conference paper at PAKDD 2020
\end{textblock*}
\vspace{-1cm}
{\let\newpage\relax\maketitle}
% \maketitle              % typeset the header of the contribution
%

\begin{abstract}
Linking job seekers with relevant jobs requires matching based on not only skills, but also personality types.  Although the Holland Code also known as RIASEC has frequently been used to group people by their suitability for six different categories of occupations, the RIASEC category labels of individual jobs are often not found in job posts. This is attributed to significant manual efforts required for assigning job posts with RIASEC labels. To cope with assigning massive number of jobs with RIASEC labels, we propose \textsc{JPLink}, a machine learning approach using the text content in job titles and job descriptions. \textsc{JPLink} exploits domain knowledge available in an occupation-specific knowledge base known as \textsc{O*NET} to improve feature representation of job posts. To incorporate relative ranking of RIASEC labels of each job, \textsc{JPLink} proposes a listwise loss function inspired by learning to rank. Both our quantitative and qualitative evaluations show that \textsc{JPLink} outperforms conventional baselines. We conduct an error analysis on \textsc{JPLink}'s predictions to show that it can uncover label errors in existing job posts. 

\end{abstract}
\section{Introduction}~\label{sec:intro}
\begin{figure*}[t]
\scriptsize
\centering
%\resizebox{\textwidth}{!}{%
\subfloat[]{%
%\begin{figure}
    %\centering
    \includegraphics[width=0.47\linewidth]{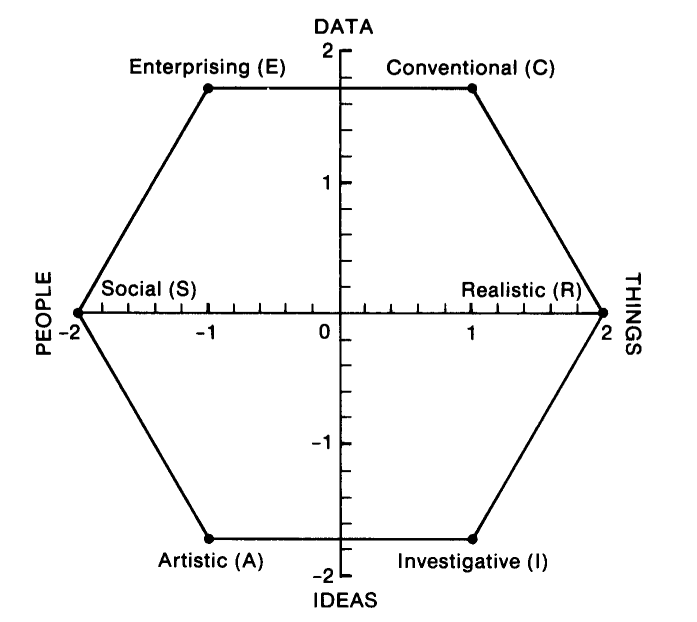}
    %\caption{Caption}
    %\label{fig:my_label}
%\end{figure}
%
\label{fig1:a}
}\hspace{1em}
\subfloat[]{%
\scriptsize
\centering
\begin{tabular}[b]{c|c}
        Occupations& Jobs\\
        \hline
        \multirow{5}{*}{Lawyer}& Attorney at Law\\
        & Lawyer\\
        & General Counsel\\
        & Assistant Counsel\\
        & Associate Attorney\\
        \hline
        \multirow{5}{*}{\shortstack{Web\\Programmer}}&Designer\\
        & Webmaster\\
        & Web Architect\\
        & Web Designer\\
        & Web Developer\\
        \hline
        \multirow{5}{*}{Librarian}&Library Director\\
        & Children's Librarian\\
        & Library Media Specialist\\
        & Catalog Librarian \\
        & Reference Librarian\\
        \hline
\end{tabular}
\label{fig1:b}
}%
%}%
\caption{(a) The Hexagonal Model of Holland's Vocational Interest Types (source:~\cite{prediger_1981}); and (b) a few examples for jobs and occupations (source: \href{https://www.onetonline.org/}{O*NET})} \label{fig:results2}
\vspace{-5.5mm}
\end{figure*}

\textbf{Motivation.}  Job profiling refers to uncovering important characteristics of jobs for generating useful insights about job trends and for matching jobs with talents. 
%In this paper, we focus on learning for each job the personality profiles of people who suit the job. 
According to Holland's theory~\cite{holland1997making}, each occupation (or applicant) can be assigned 1 to 3 out of 6 personality types characterizing different personality types. These personality types are assigned 
%different Holland Codes also known as 
RIASEC labels: \textsc{Realistic} (R), \textsc{Investigative} (I), \textsc{Artistic} (A), \textsc{Social} (S), \textsc{Enterprising} (E) and \textsc{Conventional} (C) (see Figure~\ref{fig1:a}).  Ideally, one should match people with jobs based on personality types, assuming that all other job criteria (e.g., skills, abilities, knowledge, etc.) are already met. For example, doctor, researcher, and lawyer are jobs ideal for people with \textsc{investigative} personality type, while photographer, musician, and architect are jobs ideal for people with \textsc{artistic} personality type.

Nevertheless, RIASEC labels are usually not found in the job descriptions. Experts in the past attempted to focus on manually performing personality type annotations at the occupation level as each occupation represents a set of jobs involving similar job tasks (see Figure~\ref{fig1:b} for examples). Such an approach assumes that jobs of the same occupation share the same personality types. This assumption does not always work well when people are often expected to be recommended specific jobs instead of occupations. Also, the manual approach does not timely profile new occupations which are expected to emerge at faster pace due to recent technology disruptions. 

\textbf{Objectives. }We aim to determine the personality types of a large collection of job posts. In this task, one has to address a few research challenges: (a) limited labeled data; (b) noisy word semantics; and (c) ranked personality types. 

First, there are very limited labeled data available for training and evaluation. One can certainly find pre-existing occupations labeled with personality types but not jobs. To the best of our knowledge, the number of occupations is usually at the scale of less than 1500, much smaller than the millions of job posts available. In this work, we use the labeled data available in O*NET at the occupation level, which consists of 974 different occupations with RIASEC labels. Without labeled jobs as ground truth, we have to use labeled occupations in some distant supervision solution approach.

Second, the description for the same job can be vastly different due to different word choices, job scopes and requirements. One has to accommodate these differences in developing an accurate prediction model. Although there are previous attempts to use pre-trained word embedding to profile job posts~\cite{zhu2017document}, domain-specific words are not accurately represented in pre-trained word embeddings as they often do not carry their domain-specific meanings in a general corpus. For instance, the word \textit{spark} in a software developer job mostly refers to a cluster computing framework, although it means \textit{"an emission of fire or electricity"} in a general corpus. 

Finally, up to three personality types can be associated with a job post and they are ranked. We thus require the prediction model to recover the ranking.

\textbf{Contributions.}
To address these challenges, we propose \textsc{JPLink}, a framework to profile jobs with their personality types, which: (1) jointly learns domain-specific word and occupation representations using knowledge available in O*NET. To the best of knowledge, this is the first work to learn word representations specific to occupations and jobs; (2) incorporates a novel supervised approach to assign RIASEC labels to occupations and job posts using their text context, which considers the inter-correlations among RIASEC dimensions. \textsc{JPLink} outperforms the conventional baselines by 4.86\% and it yields a high ranking accuracy measured by NDCG (=0.949); and (3) predicts the personality type labels for a set of job posts, extracted from Singapore's JobsBank\footnote{A central repository of job posts in Singapore \href{https://www.jobsbank.gov.sg/}{https://www.jobsbank.gov.sg/}}, for which weak labels are assigned using a distant supervision approach. Our error analysis shows \textsc{JPLink} can effectively overcome the imperfections caused by the limitation of assuming all jobs of the same occupation share the identical personality type. 

\section{Related Work}
\label{sec:related}

%\textbf{Job Profiling Research.} 
Personality types have been extensively studied using empirical evidence. The study in~\cite{prediger1982dimensions} provides empirical evidence of the personality type dimensions underlying the hexagon of Hollands' theory and it shows that individuals' personality types characterize the tasks they prefer to perform in jobs. \cite{prediger1992locating} introduces a procedure to associate occupations with Holland's hexagon (\textit{Hexagon Congruence Index}) using the personality type scores and empirically shows that \textit{Hexagon Congruence Index} provides a basis for a new index of congruence (e.g., person-occupation, occupation-occupation), by which similar occupations and user profiles can be identified. 

Nevertheless, there are very little efforts on assigning personality types to jobs as opposed to occupations. To the best of our knowledge, 
%all the previous efforts have focused on assigning personality types to occupations each of which represents a class of similar jobs. The proposed
there are three types of methods for occupational level personality type determination: \textit{(a)  Incumbent Method}, in which the personality type of an occupation is the average of the personality types of a representative sample of workers taking jobs of that occupation (based on the Holland's idea that  people sharing the same occupation represent the occupation).  This method requires the personality types of many individuals and is therefore very costly (i.e., less practical); \textit{(b) Empirical Method}, which uses occupational analysis data (collected using \textit{Incumbent Method}) to develop classifiers to predict the personality type of occupations; and \textit{(c) Judgment Method}, which involves trained experts making direct personality type assignment to occupations. An example for such an approach is O*NET interest profiler~\cite{rounds1999development}. It constructs initial personality type of occupations based on discriminant functions derived from the ratings of Occupational Units~\cite{rounds1999development}. Judgment method is then performed to fine-tune the constructed personality profiles using expert knowledge. Usually, human judges are trained to determine occupations' personality profiles after looking at attributes of occupations such as title, description, and job tasks. Hence, such an approach requires a considerable human effort. It is also difficult to extend this kind of approach for new emerging occupations and jobs. %Also, the classification of interest profiles based on occupational analysis data might be unreliable due to dated occupational data~\cite{gottfredson1996dictionary}. 
Almost all these previous efforts share the assumption of all job posts of the same occupation share the same personality type.  This differentiates our research from them.   

\section{Dataset Construction}\label{sec:dataset}

\begin{table}[t]
\scriptsize
\centering
\caption{Descriptive Statistics of O*NET knowldege base}\label{tab:actual_corr}
\begin{tabular}{l|c|c|c|c|c|c||c|}
\cline{2-8}
                                    & \multicolumn{6}{c||}{Actual Correlations}                                                                & \multirow{2}{*}{\shortstack{\% of \\Proportions}}\\ \cline{2-7} 
                                    & C & E & I    & S           & R        & A         &             \\ \hline %\hline
\multicolumn{1}{|l|}{Conventional (C)}  & 1.00            & 0.21        & \textbf{-0.28} & \textbf{-0.34} & -0.05          & \textbf{-0.50} &          23.76\%   \\ \hline
\multicolumn{1}{|l|}{Enterprising (E)}  &              & 1.00            & \textbf{-0.46} & 0.15           & \textbf{-0.51} & -0.06          &             14.75\% \\ \hline
\multicolumn{1}{|l|}{Investigative (I)} &              &              & 1.00                & -0.09          & -0.17          & 0.11            &            16.01\%  \\ \hline
\multicolumn{1}{|l|}{Social (S)}        &              &              &                  & 1.00                & \textbf{-0.60} & \textbf{0.24}  &             11.16\% \\ \hline
\multicolumn{1}{|l|}{Realistic (R)}     &              &              &                  &                  & 1.00                & \textbf{-0.39} &             28.18\% \\ \hline
\multicolumn{1}{|l|}{Artistic (A)}      &              &              &                  &                  &                  & 1.00                &            6.13\%  \\ \hline
\end{tabular}
\end{table}

\subsection{Occupation-Specific Knowledge Base Extraction} 

For the purpose of learning domain-specific word embedding as well as for evaluation of interest profile prediction, we crawled the \textsc{O*NET} occupation knowledge base which covers 1110 different occupations and their RIASEC profiles, which is publicly available at \href{https://www.onetonline.org/}{https://www.onetonline.org/} %\href{https://drive.google.com/open?id=1KBChBHFUQYzmUi6FbcS9CLYGGbvRlia-}\texttt{https://drive.google.com/1KBCh}. 
Each occupation has a profile consisting of six numerical scores in the range $[0,100]$, one for each RIASEC dimension. The dimensions with scores more than 50 are known as the \textit{interest codes} of the occupation. For example, an occupation with profile of (R=80, I=40, A=20, S=60, E=10, C=70) will be assigned the Holland codes R, C and S.  The rightmost column in Table~\ref{tab:actual_corr} shows the RIASEC distribution for the 1110 occupations found in O*NET. %This distribution suggests that realistic (R) and conventional (C) are the most popular dimensions among occupations, while artistic (A) is the least popular dimension.  
Table~\ref{tab:actual_corr} depicts the Spearman's correlation between personality type dimensions. We observe that %more than $50\%$ of occupations have Realistic dimension appeared within their interest code, and it is nearly $50\%$ for Conventional dimension. But only a few occupations ($\approx 13\%$) are categorized under Artistic. In addition, we can observe 
the correlations among dimension is consistent with Holland's Hexagonal Model (Figure ~\ref{fig1:a}). The opposite and consecutive dimensions in Holland's Hexagon have significant negative correlations (e.g., Realistic vs Social, Enterprising vs Investigative, and Artistic vs Conventional) and positive correlations (e.g., Artistic vs Social) respectively.  In other words, jobs for realistic people may not suit social people, jobs for enterprising people may not suit investigative people, etc..  On the other hand, artistic people may be able to take on social jobs. 

O*NET also identifies similar occupations for each occupation. We thus construct a network of similar occupations known as \textit{occupation network} and measure homophily of RIASEC labels in \textit{occupation network} using \textit{Affinity}~\cite{mislove2010you} measure, which is defined as the ratio between the observed fraction of links between interest dimension sharing occupations in the network, and the expected fraction of links between interest dimension sharing occupations. Here, we assign each occupation in the network with its highest-scored RIASEC dimension. The \textit{occupation network} reports an \textit{Affinity} of $2.82\text{ }(\gg 1)$, implying that occupations connected by a link in the network are 2.82 times more likely to share the similar dimension than that between any two random occupations. Such a strong homophily property demonstrates the importance of taking advantage of O*NET occupation network information for RIASEC prediction.
\vspace{-2mm}

\subsection{Extraction of Job Posts from Singapore's JobsBank}\label{subsec:jobpost}

To quantitatively evaluate \textsc{JPLink}, we crawled a set of 217,874 job posts from Singapore's Jobsbank posted during the period from September 2017 to December 2018. Each job post consists of fields such as job title, skill description, and SSOC (Singapore Standard Occupational Classification)\footnote{https://www.singstat.gov.sg/standards/standards-and-classifications/ssoc} category. These job posts however do not come with any personality type  profile.

{\bf Weakly Labelling Personality Type Profile of Job Posts.}
%We adopt the following distant supervision approach to recover the interest profile labels of job posts. 
As mentioned in Section~\ref{sec:related}, personality type profiles are only available at the occupation level, not at the job level. While each job post has an SSOC code corresponding to some occupation, the code is not associated with RIASEC profiles. To derive the latter, we propose a distant supervised approach to map SSOC occupations to most similar O*NET occupations (for which personality types are assigned). Since there is no direct mapping between SSOC and SOC, we first use a mapping table\footnote{\href{https://www.singstat.gov.sg/-/media/files/standards_and_classifications/occupational_classification/ssoc2015-v2018-isco-08-correspondence.xlsx}{https://www.singstat.gov.sg/ssoc2015-v2018-isco-08-correspondence.xlsx}} matching SSOC occupation codes with ESCO occupation codes which are the occupation codes standardized in the EU region. Subsequently, we determine the most similar SOC occupations for each ESCO occupation using another mapping table from Bureau of Labor Statistics of USA \footnote{\href{https://www.bls.gov/soc/ISCO_SOC_Crosswalk.xls}{https://www.bls.gov/soc/ISCO\_SOC\_Crosswalk.xls}}. Finally, the RIASEC profile of an SSOC occupation is defined as the average of its similar SOC occupations' profiles. Following the assumption that jobs have RIASEC profiles similar to their occupations, we assign weak interest profile labels to the job posts. In this way, we are able to map 96.71\% of SSOC occupations to 96.57\% of ESCO occupations, and finally 75.23\% of SOC occupations. This amounts to 171,946 job posts assigned with weak interest profile labels, which are used in Section~\ref{sec:results} to train \textsc{JPLink} and to evaluate the prediction accuracy. 
\vspace{-2mm}

\section{RIASEC Profile Prediction Problem}\label{sec:problem}

We define the RIASEC profile prediction problem to consist of (a) learning domain-specific representations for words and occupations; and (b) prediction of personality profiles (RIASEC dimensions) for occupations and job posts. % The notations defined in this section are used in the rest of the paper.

Formally, let $O = \{o_1, o_2, ..., o_n\}$ is the 1110 ($=n$) occupations available in O*NET. Each occupation $o \in O$ is a tuple $\langle W_{o}, E_{o}, y_{o} \rangle$, where $W_{o}$ is a sequence of words describing the tasks, job activities, and other aspects of occupation $o$, and $E_{o}$ is the set of similar occupations to $o$. The personality profile of $o$ is denoted as $y_o$ a vector with elements $y_o^d \in \mathbb{R}^{[0,100]}$ which is the score of the $d^{th}$ dimension, where $d \in \{R,I,A,S,E,C\}$ represents the personality type dimensions. % ``conventional'', ``enterprising'', ``investigative'', ``social'', ``realistic'', and ``artistic'' respectively.
% The function $b:y_o^d  \rightarrow \{0,1\}$ defines the binary label (interest code) assignment for $d^{th}$ RIASEC dimensions of $o$, where $b(y_o^d)= min(\lfloor y_o^d/50 \rfloor,1)$ and $\lfloor . \rfloor$ is the floor operation.  {\color{red} Where do we use this binary label?}

% \begin{equation}
%     b(y_o^d) = 
%     \left\{\begin{matrix}
%  1 & \quad if\  y_o^d \geq 50 \\ 
%  0 & \quad otherwise
% \end{matrix}\right.
% \end{equation}

\textbf{Learning domain-specific word and occupation representations.} 
The goal of this task is to learn mapping function $f: w|o \rightarrow  \mathbb{R} ^{1\times k}\text{ }\text{for } o \in O, w \in W$, where $W = \bigcup_{o} W_{o}$ and $k$ is the dimension of the embedding space. $W_o$ denotes the set of words in occupation $o$.% and $k \ll |O| + |W|$. The mapping function $f$ can be initialized using pre-trained word representations $\hat{f}$, which is learned using a large general corpus.%\footnote{https://github.com/mmihaltz/word2vec-GoogleNews-vectors}.

\textbf{Prediction of RIASEC/personality type profiles for occupations and job posts.}
Here, our goal is to predict personality type profile $\hat{y}_{o}$ for a given $o \in O$ or $\hat{y}_j$ for a given job post $j \in J$, where $J$ is the set of job posts. A job post $j$ is a tuple $\langle W_j^t, W_j^d \rangle$, where both $W_j^t$ and $W_j^d$ are word sequences to represent $j$'s title and description respectively.

\begin{figure}[t]
    \centering
    \includegraphics[width=0.90\linewidth]{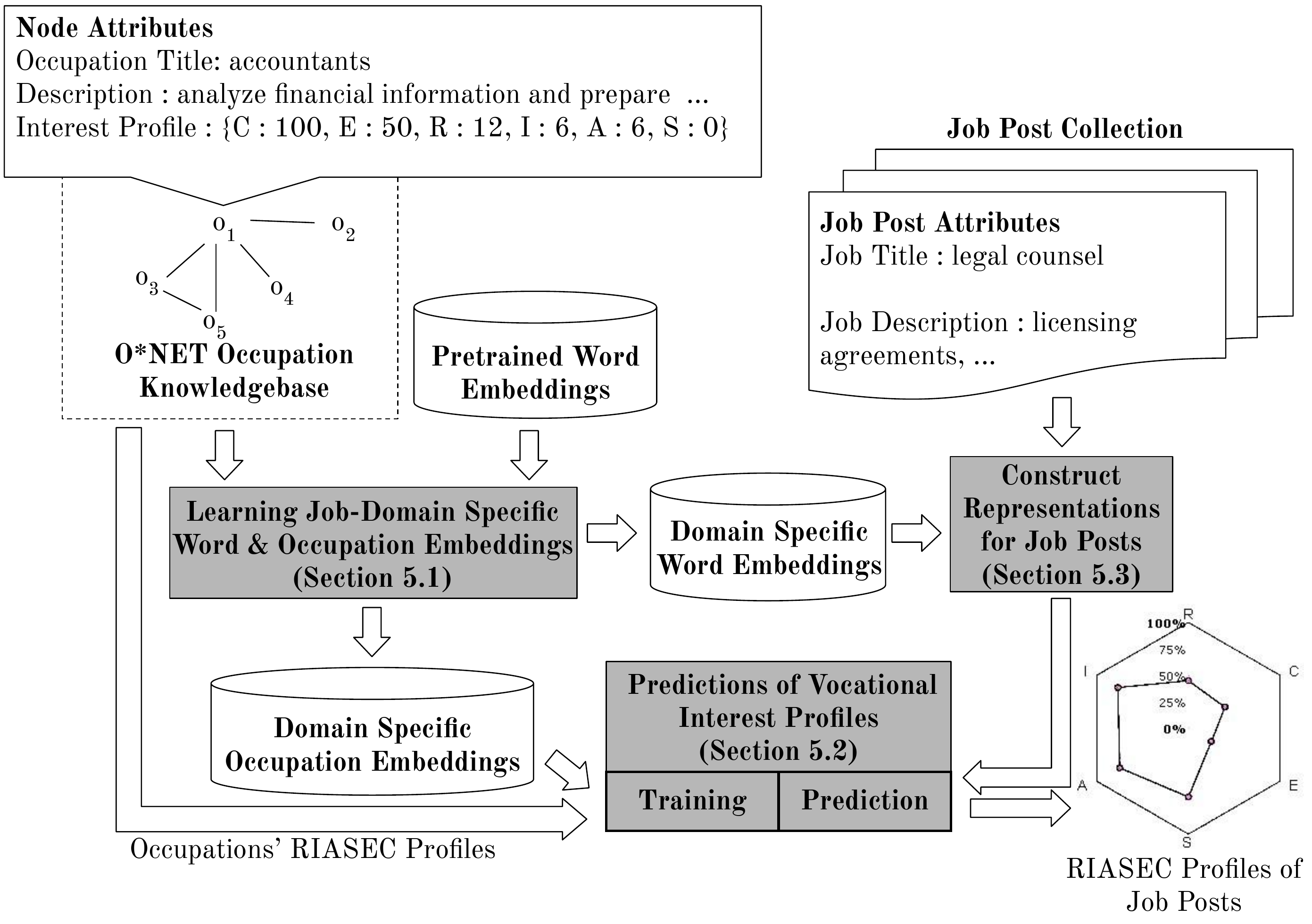}
    \caption{\textsc{JPLink} Framework}
    \label{fig:framework}
\vspace{-6mm}
\end{figure}

\section{\textsc{JPLink}}\label{sec:model}
Our proposed \textsc{JPLink} framework consists of three parts as depicted in Figure~\ref{fig:framework}: (1) learning job-domain specific embedding for words and occupations (\textsc{JPLink-Emb}); (2) prediction of personality type profiles for occupations and jobs (\textsc{JPLink-Pred}); and (3) representation construction for new occupations or job posts.

\subsection{Learning Job-Domain Specific Embedding for Words and Occupations (\textsc{JPLink-Emb})}\label{subsec:model1}
\textsc{JPLink-Emb} is motivated by skip-gram model~\cite{mikolov2013distributed}, a popular language model in NLP. For a given set of sentences, the skip-gram model loops on the words in each sentence and uses the current word to predict its neighbors. Formally, for a given sentence $w_1, w_2, w_3, . . . , w_T$, the skip-gram model maximizes:\\
$    \frac{1}{T}\sum_{t=1}^{T}\sum_{-c \leq j \leq c} log\text{ }p(w_{t+j}|{w_t}) 
$
%\begin{equation}
%\label{eq:skipgram}
%    \frac{1}{T}\sum_{t=1}^{T}\sum_{-c \leq j \leq c} log\text{ }p(w_{t+j}|{w_t}) 
%\end{equation}
where $c$ is the window size around the center word $w_t$. By doing so, representations of the words with similar context will be closed to each other in the embedding space (second order proximity is preserved). Generally, the skip-gram model is trained on a large general text corpora (i.e., Wikipedia and Google News) which do not carry any occupation-specific knowledge. The trivial way of constructing representations for multi-word entities is taking the mean representation of the words in the particular entity (e.g., the representation of ``computer programmer'' is the mean of the representations of ``computer'' and ``programmer''). We refer this approach as \textsc{PretrainedEmb} for the rest of this paper.

\textbf{Knowledge Graph Construction. }In our model, a knowledge graph $G$ is constructed, in which both $w \in W$  and $o \in O$ are nodes. We create edges between two occupations by connecting each $o\in O$ to the similar occupations in $E_o$. For each $o \in O$, a set of discriminative words are selected from $W_o$ and these are connected to the $o$'s node. This derives a graph $G$ with useful first-order context to represent each node (word or occupation). 

We perform random walks around each node in $G$ and then the skip-gram model is used to learn representations for nodes, considering random walk sequences are analogous to sentences in a language. Due to the importance of first-order context, we perform breadth-first random walks with random restarts (inspired by node2vec~\cite{grover2016node2vec}). We observe that such a procedure provides the relevant context for each node and outperforms conventional network representation learning techniques (e.g., deepwalk \cite{perozzi2014deepwalk} and LINE \cite{tang2015line}) for this particular task. 

In this work, the discriminative words are determined empirically by their normalized document frequency $< 10\%$. We leave the exploration of other threshold settings and the embedding schemes for knowledge graph to future work.

\subsection{Predictions of Personality Type Profiles for Occupations and Jobs (\textsc{JPLink-Pred})}\label{subsec:model2}

We formulate the predicted personality type profile ($\hat{y}_o \in \mathbb{R} ^{1\times 6}$) of an occupation $o$ as a linear projection $A$ of input features followed by a bias offset $b$ and a softmax activation function. Formally, $\hat{y}_o = softmax(A*f(o)^T + b)$, where $f(o)$ denotes the embedding of $o$ while $A_{6\times k}$ and $b_{1 \times 6}$ are the trainable linear projection matrix and bias offsets respectively.

% \begin{equation}\label{eq:model_eq}
%     \hat{y}_o = softmax(A*f(o)^T + b)
% \end{equation}
% \begin{equation}\label{eq:model_eq}
% \begin{split}
%     \hat{y}_o = softmax(A*f(o)^T + b)
%     % \phi (o)& = softmax(A*f(o)^T + b)\\
%     %\hat{y}_o& = \frac{\exp(\phi(o))}{\sum_{d \in \{R, I, A, S, E, C\}} \exp(\phi(o)^d)}\\
% \end{split}
% \end{equation}

 As we defined in Section~\ref{sec:problem}, the objective of this prediction model is to produce correct ranking of the personality type dimensions for a given $o$. This motivates us to propose an approach to predict the personality type profiles by learning to rank~\cite{burges2005learning,cao2007learning}. We introduce a listwise loss function for learning, which consider the relative ranking information of all RIASEC dimensions together. We adopt ListNet~\cite{cao2007learning}, a listwise learning to rank method, to compute the target top one probability of dimension $d$ for a given occupation $o$ as,
 
\begin{equation}\label{eq:listnet}
    P_{d}(y_o) = \frac{\exp{y_o^{d}}}{\sum_{d^{'} \in \{R, I, A, S, E, C\}}\exp{y_o^{d^{'}}}}
\end{equation}
$\hat{y}_{o}$ is used as the corresponding modeled top one probability. Then, \textsc{JPLink-Pred} optimizes the following loss function using SGD,

\begin{equation}
    L(y_o, \hat{y}_{o}) = \sum_{d \in \{R, I, A, S, E, C\}} P_d(y_o)*\log(\hat{y}_{o}^d) + (1 - P_d(y_o))*\log(1-\hat{y}_{o}^d)  
\end{equation}

\subsection{Construct Representations of New Occupations or Jobs}\label{subsec:model3}

Once the prediction model is learned, we can apply the model to predict personality type profiles of jobs. Before that, we construct the representations of a new job. Given a job $j$, we use words appearing in the job title $W_j^t$ and job description $W_j^d$ to generate representation $f(j)$ as shown in Equation~\ref{eq:job_post}.  The representation for a new occupation can be constructed in the same way.  
\begin{equation}\label{eq:job_post}
    \begin{split}
        % f(W_j^t) &= \frac{\sum_{w \in W_j^t}f(w)}{\sum_{w \in W_{j}^t}1}\\
        % f(W_j^d) &= \frac{\sum_{w \in W_j^d}f(w)}{\sum_{w \in W_{j}^d}1}\\
        f(j) &= \beta * f(W_j^t) + (1 - \beta) * f(W_j^d)
    \end{split}    
\end{equation}
where $f(W_j^t) = \sum_{w \in W_j^t}f(w)/\sum_{w \in W_{j}^t}1$ and $f(W_j^d) = \sum_{w \in W_j^d}f(w)/\sum_{w \in W_{j}^d}1$ denote the representations constructed for the job (or occupation) title and job (or occupation) description respectively. The parameter $\beta$ controls the importance given to title and description in the final representation. The optimal value of $\beta$ is determined using a grid search. %{\color{} How is $\beta$ determined in the experiments?} When $\beta=0$, the representation will reduce to the average of word embeddings of words in the job (or occupation) description which has been used in some previous works~\cite{zhu2017document}.  

\section{Evaluation}
\label{sec:results}
\vspace{-2mm}

\textbf{Baselines. }
We compare \textsc{JPLink-Emb} with two baselines: (1) \textsc{PretrainedEmb}, which adopts the pretrained word embeddings\footnote{\href{https://github.com/mmihaltz/word2vec-GoogleNews-vectors}{https://github.com/mmihaltz/word2vec-GoogleNews-vectors}} as introduced in Section~\ref{subsec:model1}; and (2) \textsc{Wikipedia2Vec}, which adopts the model proposed in~\cite{yamada2016joint} to jointly learn embedding for words and multi-word entities, considering occupations as entities.

We compare \textsc{JPLink-Pred} with two well known baselines: (1) \textsc{Point}, which adopts Logistic Regression classifier with binary cross entropy loss. \textsc{Point} ignores the relative ranking of the RIASEC dimensions; and (2) \textsc{Pair}, which adopts the pairwise learning to rank loss function proposed in \textsc{RankNet} \cite{burges2005learning}. \textsc{RankNet} considers the relative ranking information between pairs of output dimensions.

\textbf{Measuring Performance. } 
As claimed by~\cite{gati1998using}, occupations should be characterized by a variable size set of interest dimensions. Hence the prediction engine should be capable to produce the correct ranking of RIASEC dimension for a given occupation $o$ or job $j$. We thus measure the performance of RIASEC dimension prediction using Normalized Discounted Cumulative Gain (NDCG), which is commonly used to evaluate the performance of rankers in a multi-graded relevance setting. To calculate NDCG, each RIASEC dimension ($d \in \{R,I,A,S,E,C\}$) is assigned a relevance score $R_{d}(o)$ using the ground truth score ($y_o^{d}$) taken from O*NET for $o$ as $R_{d}(o) = \lfloor y_o^{d}/20 \rfloor$, where $\lfloor . \rfloor$ denotes the conventional floor operator. Suppose a method ranks the interest dimensions of an occupation $o$ as $\hat{\sigma}_o$ such that $\hat{\sigma}_o(l)$ ($1 \leq l \leq 6$) returns the $l^{th}$ ranked interest dimension.  
We then define NDCG to be:
% \begin{equation}\label{eq:ndcg}
% \begin{split}
% %    NDCG(\hat{\sigma}(o_i), R(o_i)) &= \frac{DCG(\hat{\sigma}(o_i), R(o_i))}{\max_{\sigma(o_i)}DCG(\sigma(o_i), R(o_i))}\\
% %    DCG(\sigma(o_i), R(o_i)) &= \sum_{d = 1}^{6} \frac{2^{R_d(o_i)}-1}{\log_2(1 + \sigma_d(o_i))}\\
%     NDCG(\sigma_o) &= \frac{DCG(\sigma_o)}{\max_{\sigma'_o}DCG(\sigma'_o)}\\ 
%     DCG(\sigma_o) &= \sum_{l = 1}^{6} \frac{2^{R_{\sigma_o(l)}(o)}-1}{\log_2(1 + l)}\\
% \end{split}    
% \end{equation}
\begin{equation}\label{eq:ndcg}
%    NDCG(\hat{\sigma}(o_i), R(o_i)) &= \frac{DCG(\hat{\sigma}(o_i), R(o_i))}{\max_{\sigma(o_i)}DCG(\sigma(o_i), R(o_i))}\\
%    DCG(\sigma(o_i), R(o_i)) &= \sum_{d = 1}^{6} \frac{2^{R_d(o_i)}-1}{\log_2(1 + \sigma_d(o_i))}\\
    NDCG(\sigma_o) = \frac{DCG(\sigma_o)}{\max_{\sigma'_o}DCG(\sigma'_o)}\quad\quad
    DCG(\sigma_o) = \sum_{l = 1}^{6} \frac{2^{R_{\sigma_o(l)}(o)}-1}{\log_2(1 + l)}
\end{equation}
NDCG for job $j$ can be similarly defined by replacing $o$ by $j$, and $\sigma_o$ by $\sigma_j$.
%\vspace{-2mm}
\subsection{Evaluation of the Job-Domain Specific Embeddings}

\begin{table}
\scriptsize
\caption {5 most similar occupations and words for a set of occupations, as induced by different embeddings.} \label{tab:qual_analysis}
\noindent\makebox[\textwidth]{%
\begin{tabular}{c|c|c|c|c|c|c|}
\cline{2-7}
\textbf{}                                                                                                     & \multicolumn{3}{c|}{\textbf{5 most similar occupations}}                                                                                                                                                                                                                                                                                                                                           & \multicolumn{3}{c|}{\textbf{5 most similar words}}                                                                                                               \\ \hline
\multicolumn{1}{|c|}{\textbf{\begin{tabular}[c]{@{}c@{}}Target \\ Occupation\end{tabular}}}                   & \textbf{\textsc{PretrainedEmb}}                                                                                                       & \textbf{\textsc{JPLink-Emb}}                                               & \textbf{\textsc{Wikipedia2Vec}}                                                                               & \textbf{\textsc{PretrainedEmb}} & \textbf{\textsc{JPLink-Emb}} & \textbf{\textsc{Wikipedia2Vec}} \\ \hline\hline
\multicolumn{1}{|c|}{\multirow{5}{*}{\textbf{\begin{tabular}[c]{@{}c@{}}financial \\ managers\end{tabular}}}} & \begin{tabular}[c]{@{}c@{}}investment fund \\ managers\end{tabular}                                                     & \begin{tabular}[c]{@{}c@{}}financial \\specialists, \\ all other  \end{tabular}                                                                                                & \begin{tabular}[c]{@{}c@{}}auditors\end{tabular}                                                                                      & managers          & investing                                                            & brokerage                                                             \\ \cline{2-7} 
\multicolumn{1}{|c|}{}                                                                                        & \begin{tabular}[c]{@{}c@{}}security \\ managers\end{tabular}                                                                                                       & auditors                                 & \begin{tabular}[c]{@{}c@{}}sales agents, \\ financial services\end{tabular}                                                                        & executives        & banking                                                              & issuers                                                               \\ \cline{2-7} 
\multicolumn{1}{|c|}{}                                                                                        & \begin{tabular}[c]{@{}c@{}}marketing \\managers\end{tabular}                                                                                                      & \begin{tabular}[c]{@{}c@{}}financial \\managers, \\branch or\\ department\end{tabular}                                                                                                            & \begin{tabular}[c]{@{}c@{}}transportation, \\storage, and \\distribution \\managers\end{tabular}                                                      & bankers           & branch                                                               & banking                                                               \\ \cline{2-7} 
\multicolumn{1}{|c|}{}                                                                                        & \begin{tabular}[c]{@{}c@{}}logistics \\managers \end{tabular}                                                                                                     & \begin{tabular}[c]{@{}c@{}}treasurers and \\ controllers\end{tabular}                                         & \begin{tabular}[c]{@{}c@{}}financial \\managers, branch\\or department\end{tabular}                                                                & investors         & securities                                                           & intermediary                                                          \\ \cline{2-7} 
\multicolumn{1}{|c|}{}                                                                                        & \begin{tabular}[c]{@{}c@{}}administrative \\services\\ managers\end{tabular}                                             & loan officers                                                                                                       & \begin{tabular}[c]{@{}c@{}}treasurers and \\ controllers\end{tabular}                                                                              & investment        & accounting                                                           & entities                                                              \\ \hline\hline
\multicolumn{1}{|c|}{\multirow{5}{*}{\textbf{\begin{tabular}[c]{@{}c@{}}technical \\ writers\end{tabular}}}}  & \begin{tabular}[c]{@{}c@{}}poets, lyricists\\and creative\\writers\end{tabular}                                        & \begin{tabular}[c]{@{}c@{}}poets, lyricists\\and creative\\writers\end{tabular}                                    & \begin{tabular}[c]{@{}c@{}}broadcast news \\ analysts\end{tabular}                                                                                 & writers           & delineate                                                            & detail                                                                \\ \cline{2-7} 
\multicolumn{1}{|c|}{}                                                                                        & \begin{tabular}[c]{@{}c@{}}writers and \\authors\end{tabular}                                                                                                         & \begin{tabular}[c]{@{}c@{}}writers and \\authors\end{tabular}                                                                                                 & \begin{tabular}[c]{@{}c@{}}interpreters and \\ translators\end{tabular}                                                                            & technical         & published                                                            & planes                                                                \\ \cline{2-7} 
\multicolumn{1}{|c|}{}                                                                                        & \begin{tabular}[c]{@{}c@{}}technical \\ directors/\\managers\end{tabular}                                                 & \begin{tabular}[c]{@{}c@{}}reporters and \\ correspondents\end{tabular}                                             & \begin{tabular}[c]{@{}c@{}}poets, lyricists \\and creative\\writers\end{tabular}                                                                   & editors           & mockups                                                              & freelancers                                                           \\ \cline{2-7} 
\multicolumn{1}{|c|}{}                                                                                        & copy writers                                                                                                            & editors                                                                                                             & editors                                                                                                                                            & programmers       & illustrate                                                           & published                                                             \\ \cline{2-7} 
\multicolumn{1}{|c|}{}                                                                                        & \begin{tabular}[c]{@{}c@{}}gaming and \\sports book\\writers and\\runners\end{tabular}                                   & copy writers                                                                                                        & \begin{tabular}[c]{@{}c@{}}proofreaders and\\ copy markers\end{tabular}                                                                            & writing           & terminology                                                          & publications                                                          \\ \hline\hline
\multicolumn{1}{|c|}{\multirow{5}{*}{\textbf{\begin{tabular}[c]{@{}c@{}}health \\ educators\end{tabular}}}}   & \begin{tabular}[c]{@{}c@{}}self-enrichment \\ education \\teachers\end{tabular}                                           & \begin{tabular}[c]{@{}c@{}}rehabilitation \\counselors\end{tabular}                                                                                           & \begin{tabular}[c]{@{}c@{}}speech-language \\ pathology \\assistants\end{tabular}                                                                    & educators         & lifestyles                                                           & impart                                                                \\ \cline{2-7} 
\multicolumn{1}{|c|}{}                                                                                        & \begin{tabular}[c]{@{}c@{}}community\\health workers\end{tabular}                                                     & \begin{tabular}[c]{@{}c@{}}speech-language \\ pathology\\assistants\end{tabular}                                     & \begin{tabular}[c]{@{}c@{}}rehabilitation \\ counselors\end{tabular}                                                                               & teachers          & smoking                                                              & equipping                                                             \\ \cline{2-7} 
\multicolumn{1}{|c|}{}                                                                                        & \begin{tabular}[c]{@{}c@{}}adult basic \\and secondary \\education and \\literacy teachers \\ and instructors\end{tabular}                                   & \begin{tabular}[c]{@{}c@{}}middle school \\teachers, except \\special and \\career/technical \\education\end{tabular} & \begin{tabular}[c]{@{}c@{}}low vision\\therapists, \\orientation and \\mobility specialists, \\and vision \\rehabilitation \\therapists\end{tabular} & physicians        & vaccines                                                             & nutritionally                                                         \\ \cline{2-7} 
\multicolumn{1}{|c|}{}                                                                                        & \begin{tabular}[c]{@{}c@{}}medical and \\health services\\managers\end{tabular}  & \begin{tabular}[c]{@{}c@{}}recreational \\therapists \end{tabular}                                                                                             & \begin{tabular}[c]{@{}c@{}}recreational \\therapists \end{tabular}                                                                                                                           & pediatricians     & cooperative                                                          & participation                                                         \\ \cline{2-7} 
\multicolumn{1}{|c|}{}                                                                                        & \begin{tabular}[c]{@{}c@{}}health \\specialties \\ teachers, \\postsecondary\end{tabular}                                         & \begin{tabular}[c]{@{}c@{}}substance abuse \\and behavioral \\disorder \\counselors\end{tabular}                    & \begin{tabular}[c]{@{}c@{}}substance abuse \\and behavioral \\disorder \\counselors\end{tabular}                                                   & nurses            & bulletins                                                            & cooperative                                                           \\ \hline\hline
\multicolumn{1}{|c|}{\multirow{5}{*}{\textbf{archivists}}}                                                    &     historians                                    & curators                                                                                                            & curators  & archivists        & archivists                                                           & archivists                                                            \\ \cline{2-7} 
\multicolumn{1}{|c|}{}                                                                                        & \begin{tabular}[c]{@{}c@{}}museum \\technicians and \\conservators\end{tabular}                                                                                                                & \begin{tabular}[c]{@{}c@{}}library science \\teachers,\\ postsecondary\end{tabular}                                  & \begin{tabular}[c]{@{}c@{}}library science \\ teachers, \\postsecondary\end{tabular}                                                                                                                                          & archival          & retained                                                             & curators                                                              \\ \cline{2-7} 
\multicolumn{1}{|c|}{}                                                                                        & librarians                                                                                                              & \begin{tabular}[c]{@{}c@{}}history teachers, \\ postsecondary\end{tabular}                                          & librarians                                                                                                                                         & archives          & valuable                                                             & librarians                                                            \\ \cline{2-7} 
\multicolumn{1}{|c|}{}                                                                                        & curators                                                                                                                & librarians                                                                                                          & \begin{tabular}[c]{@{}c@{}}transportation \\ inspectors\end{tabular}                                                                               & historians        & exhibiting                                                           & concertmasters                                                        \\ \cline{2-7} 
\multicolumn{1}{|c|}{}                                                                                        & \begin{tabular}[c]{@{}c@{}}anthropologists \\ and \\archeologists\end{tabular}                                            & historians                                                                                                          & editors                                                                                                                                            & librarians        & retrievable                                                          & compliant                                                             \\ \hline
\end{tabular}
}%
\end{table}

In this section, we compare different embedding techniques (i.e., \textsc{JPLink-Emb}, \textsc{PretrainedEmb} and \textsc{Wikipedia2Vec}). The size of the embedding is set to 300 for all three methods. We train both \textsc{JPLink-Emb} and \textsc{Wikipedia2Vec} using \textsc{O*NET}, which are initialized using \textsc{PretrainedEmb}. 

\textbf{Qualitative Evaluation.}  Our first evaluation is qualitative: we randomly select a few target occupations and manually inspect the 5 most similar occupations and words to the target occupations (by cosine similarity) using the three occupation/word embedding representations. The results are shown in Table~\ref{tab:qual_analysis}.% Due to the space limitations, we direct readers to~\cite{silva2020supplement} for the detailed results of the experiment.

We first discuss the most similar occupations. For almost all the target occupations, \textsc{PretrainedEmb} returns similar occupations which have overlapping words with the target occupations.  For example, the target occupation ``financial managers'' is similar to several other occupations of ``manager'' role.  The target occupation ``technical writers'' is similar to occupations containing ``technical'' and/or ``writers'' in their titles. On the other hand, \textsc{JPLink-Emb} produces similar occupations that may not have overlapping words.  For example, \textsc{JPLink-Emb} shows ``financial managers'' and ``loan officers'' are similar, ``health educators'', ``rehabilitation counselors'' and ``recreational therapists'' are similar, which are reasonable with respect to their domains. In contrast, \textsc{Wikipedia2Vec} produces some of the similar occupations that are less obvious, e.g., an occupation similar to ``technical writers'' is ``broadcast news analysts''. Also, when we compute the proportion of the 20 most similar occupations returned by each embedding scheme appearing as similar occupations of the target occupations in O*NET.
We found that only 20\% of similar occupations by \textsc{PretrainedEmb} appears in O*NET for the target occupations. In contrast, \textsc{JPLink-Emb} and \textsc{Wikipedia2Vec} show  55\% and 40\% of their 20 most similar occupations appearing in \textsc{O*NET} respectively. This shows the importance of incorporating O*NET knowledge base into the learning of occupation representations.

We next analyse the most similar words to the target occupations. \textsc{PretrainedEmb} returns either target occupation title words or their synonyms. E.g., \textit{``writers''} and \textit{``technical''} are the most similar words for technical writers, while \textit{``investors''} and \textit{``bankers''} are for financial managers.
\textsc{JPLink-Emb} and \textsc{Wikipedia2Vec}, on the other hand, returns domain-specific words associated with the occupational tasks.  E.g., \textit{``lifestyle'', ``smoking'', ``vaccines'' and ``nutritionally''} are words relevant to the tasks of the target occupation health educators. These words are useful to understand the requirement of health educator occupation. Because usually words overlapped with the title and its synonyms do not appear in job descriptions. Instead of that domain-specific information (tasks and skills) related to the jobs are appeared in the description.

\begin{table}[t]
\scriptsize
\centering
\caption{RAISEC Profile Prediction Result - NDCG@6}\label{tab:result}
\begin{tabular}{l|>{\centering}m{1.2cm}|>{\centering}m{1.2cm}|c|}
\cline{2-4}
& \multicolumn{3}{c|}{\textbf{Prediction Methods}}                                                                                                                                                        \\ \hline
\multicolumn{1}{|l|}{\textbf{Embedding Methods}} & POINT & PAIR & \textsc{JPLink-Pred} \\ \hline
\multicolumn{1}{|l|}{\textsc{PretrainedEmb}}     & 0.905                                                                               & 0.912                                                               & 0.928                                                                \\ \hline
\multicolumn{1}{|l|}{\textsc{Wikipedia2Vec}}   & 0.921                                                                               & 0.924                                                               & 0.941                                                                \\ \hline
\multicolumn{1}{|l|}{\textsc{JPLink-Emb}}   & {\bf 0.928}                                                                               & {\bf 0.934}                                                               & {\bf 0.949}                                                              \\ \hline
\end{tabular}
\vspace{-4mm}
\end{table}

\begin{table}[t]
\scriptsize
\centering
\caption{Correlations among the dimensions of the linear transformation matrix and the bias values in \textsc{JPLink}}\label{tab:model_para}
\begin{tabular}{l|c|c|c|c|c|c|c|}
\cline{2-8}
                                    & \multicolumn{6}{c|}{Correlations}              & Bias \\ \cline{2-8} 
                                    & C & E & I    & S           & R        &   A       &             \\ \hline
\multicolumn{1}{|l|}{Conventional (C)}  & 1.00            & 0.09       & -0.11          & \textbf{-0.29} & 0.08           & \textbf{-0.52} & 0.60      \\ \hline
\multicolumn{1}{|l|}{Enterprising (E)}  &              & 1.00            & \textbf{-0.55} & 0.12           & \textbf{-0.31} & \textbf{-0.23} & 0.12      \\ \hline
\multicolumn{1}{|l|}{Investigative (I)} &              &              & 1.00                & \textbf{-0.40} & 0.10           & -0.07          & 0.15      \\ \hline
\multicolumn{1}{|l|}{Social (S)}        &              &              &                  & 1.00                & \textbf{-0.52} & -0.01          & -0.63     \\ \hline
\multicolumn{1}{|l|}{Realistic (R)}     &              &              &                  &                  & 1.00                & \textbf{-0.34} & 0.65      \\ \hline
\multicolumn{1}{|l|}{Artistic (A)}      &              &              &                  &                  &                  & 1.00                & -0.89     \\ \hline
\end{tabular}
\vspace{-4mm}
\end{table}

\textbf{Quantitative Evaluation.}
To quantitatively evaluate our end-to-end framework, we report the results for the task of predicting the RIASEC labels for occupations. In this experiment, we trained the prediction models (POINT, PAIR and \textsc{JPLink-Pred}) using 50\% of 974 O*NET occupations and then the model is evaluated using the predicted RIASEC labels for the rest 50\%, using different occupation representations. As shown in Table~\ref{tab:result}, \textsc{JPLink-Emb} outperforms \textsc{PretrainedEmb} and \textsc{Wikipedia2Vec} for all the three prediction methods. \textsc{Wikipedia2Vec} is slightly better than \textsc{PretrainedEmb} in this prediction task. These results are consistent with our qualitative analysis results.  

Among the three prediction methods, POINT shows the worst performance as it does not consider the relative ranking of RIASEC dimensions. And \textsc{JPLink-Pred} consistently outperforms PAIR which only considers the relative ranking of pairs of RIASEC dimensions. This observation shows the importance of having ranking information of RIASEC dimensions in the training phase to capture significant relationships among RIASEC dimensions in Table~\ref{tab:actual_corr}. To illustrate this fact further, we further analyze the learned parameters of the \textsc{JPLink-Pred} model.  As depicted in Table~\ref{tab:model_para}, in \textsc{JPLink-Pred}, each row in the $A$ projection matrix corresponds to the linear mapping learned for a RIASEC dimension. As shown in Table~\ref{tab:model_para}, the correlations among the rows of $A$ reflect the actual correlations between the corresponding RIASEC dimensions. These correlations are consistent with those seen in Table~\ref{tab:actual_corr}.  Moreover, we also found the bias values ($b$) of our model are consistent with the actual distribution of RIASEC dimensions in O*NET. Hence, we can say that \textsc{JPLink} is capable of capturing the inherent patterns of the interest dimensions.  
\vspace{-4mm}
\subsection{Prediction of personality type profiles for Job Posts}
\begin{figure}[t]
\centering
\begin{tikzpicture}
\begin{axis}[
	height = 5.2 cm,
	width = 7.0 cm,
    xlabel={$\beta$ value},
    ylabel={NDCG@6},
    xmin=0, xmax=1,
    ymin=0.82, ymax=0.90,
    xtick={ 0, 0.2, 0.4, 0.6, 0.8, 1},
    ytick={ 0.84, 0.86, 0.88},
    legend pos=south east,
    ymajorgrids=true,
    grid style=dashed,
]
 
%[ color=blue, mark=square,]
\addplot coordinates{(0, 0.8426) (0.1, 0.8519) (0.2, 0.8598) (0.3, 0.8664)
(0.4, 0.8715) (0.5, 0.875) (0.6, 0.8766) (0.7, 0.8764)
(0.8, 0.8734) (0.9, 0.8688) (1.0, 0.8651)};
% \addplot coordinates{(0, 0.8334) (0.1, 0.8367) (0.2, 0.8432) (0.3, 0.8515)
% (0.4, 0.8576) (0.5, 0.8629) (0.6, 0.8671) (0.7, 0.8698)
% (0.8, 0.8634) (0.9, 0.8576) (1.0, 0.8549)}; 
%\addplot coordinates{(0.5, 0.5792) (2 ,0.6616) (4,0.6659) (6, 0.6722) (8 ,0.6978) (10, 0.6956) };
\legend{\emph{JPLink}}
\end{axis}
\end{tikzpicture}
\caption{Results for the prediction of job posts' interest profiles with different $\beta$ values}\label{fig:assignment}
\vspace{-4mm}
\end{figure}
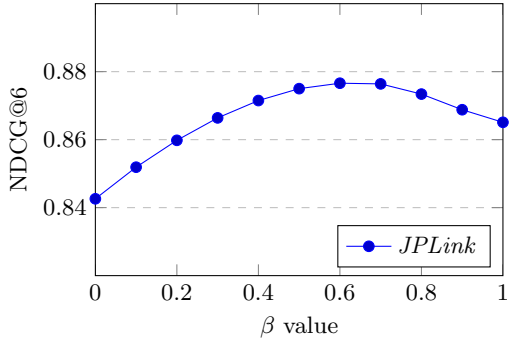

\begin{table}[t]
\scriptsize
\centering
\caption{A few job posts predicted with ``wrong'' RIASEC profile by \textsc{JPLink}}\label{tab:errors} 
\begin{tabular}{|c|c|c|c|}
\hline
                                                                                     
\begin{tabular}[c]{@{}c@{}}Assigned \\SSOC Occupation\end{tabular}& Job Title                                                            & \begin{tabular}[c]{@{}c@{}}key words in \\Job Description\end{tabular}                                                                                                                                 & \begin{tabular}[c]{@{}c@{}}Actual ($y$) and \\ Predicted ($\hat{y}$) RIASEC \\Ranking (first has \\the highest rank)\end{tabular} \\ \hline \hline
\begin{tabular}[c]{@{}c@{}}Graphic and Multimedia \\Designers and Artists\end{tabular}                       & \begin{tabular}[c]{@{}c@{}}lead full \\ stack developer\end{tabular} & \begin{tabular}[c]{@{}c@{}}widgets saas javascript \\html css community\end{tabular}                                                           & \begin{tabular}[c]{@{}c@{}}$y: \{A, R, E, I, C, S\}$\\$\hat{y}:\{C, I, R, E, S, A\} $ \end{tabular}                                                                                                               \\ \hline
\begin{tabular}[c]{@{}c@{}}Graphic and \\Multimedia Designers\\ and Artists\end{tabular}                       & \begin{tabular}[c]{@{}c@{}}singapore \\researcher  \end{tabular}                                                 & \begin{tabular}[c]{@{}c@{}}subject matter experts project\\managers executive leadership \\travel presentations oral \\independent\end{tabular} & \begin{tabular}[c]{@{}c@{}}$y: \{A, R, E, I, C, S\}$\\$\hat{y}:\{I, C, R, E, S, A\}$ \end{tabular}\\ \hline
\begin{tabular}[c]{@{}c@{}}Other Craft and \\Related Workers\end{tabular}                                                                               & line leader                                                          & coaching                                                                                                                                        & \begin{tabular}[c]{@{}c@{}}$y: \{R, E, C, A, I, S\} $\\$\hat{y}:\{E, C, S, A, R, I\}$ \end{tabular}                                                                             \\ \hline
\begin{tabular}[c]{@{}c@{}}Manufacturing \\Labourers and \\Related Workers\end{tabular}                        & on executive                                                         & \begin{tabular}[c]{@{}c@{}}online market place inventory \\ management customer service \\stocks filing enquires commerce\end{tabular}     & \begin{tabular}[c]{@{}c@{}}$y: \{R, C, E, I, A, S\}$\\$\hat{y}:\{C, E, I, S, R, A\}$ \end{tabular}                                                            \\ \hline

 \begin{tabular}[c]{@{}c@{}}Information and \\Communications\\ Technology Installers\\ and Servicers\end{tabular} & part time coach                                                      & \begin{tabular}[c]{@{}c@{}}workshops spark \\pointers learning \\funds curiosity\end{tabular}                                                     & \begin{tabular}[c]{@{}c@{}}$y: \{R, C, I, E, S, A\}$\\$\hat{y}:\{E, S, C, I, A, R\}$ \end{tabular}                                                                            \\ \hline
\end{tabular}
\end{table}

Here, we analyze the predictive power of our model trained using occupation data to predict the personality type profiles of 171,946 job posts with weak RIASEC profiles (see Section~\ref{sec:dataset}). To derive the representation of these job posts, we use the approach introduced in Equation~\ref{eq:job_post}. We observe that \textsc{JPLink-Pred} model with \textsc{JPLink-Emb} embeddings consistently gives the best performance for all different $\beta$ settings. The $\beta$ values in the range of $[0.6,0.7]$ give the best performance for \textsc{JPLink} as shown in Figure~\ref{fig:assignment}. This means that job title should be given more weight ($60\%$) when constructing the representation of job posts, which is consistent with the results in~\cite{zhu2017document}.

\textbf{Error Analysis.}
In the above results, we use the weak labels derived by a distant supervision approach for the evaluation. There might be some imperfections in these labels. In addition, we use the SSOC occupation label assigned to job posts from Singapore Jobsbank in the weakly labeling approach (see Section~\ref{sec:dataset}), which might not be perfectly accurate too. To identify whether these imperfections adversely affect our evaluation, we manually inspect a few job posts that are wrongly predicted from our model. Table~\ref{tab:errors} lists a few cases where \textsc{JPLink} wrongly predicts labels. In these example cases, job titles and descriptions do not tally with the corresponding occupations (assigned by the current system). For instance, the job title \textit{lead full stack developer} is assigned to the occupation category, "Graphics and Multimedia Designers and Artists", instead of an occupation related to software development (e.g., Software Developer). \textsc{JPLink} actually predicts a personality type profile appropriate to the occupations like Software Developer ($\{I, R, C, E, A, S\}$). Similarly, the predicted interest profiles for other example cases are reasonable with respect to their descriptions and titles. These results further signify the potential of this work to replace the current labor-intensive manual approach to profile job posts. 
\vspace{-2mm}
\section{Conclusion}\label{sec:conclusion}
\vspace{-2mm}
In this paper, we proposed \textsc{JPLink}, a framework to automate the profiling of jobs with their interest profiles. \textsc{JPLink}: (1) explored the domain-specific knowledge available in O*NET to improve existing word and occupation representations, which are subsequently used as input features to assign corresponding interest profiles; and (2) proposed a novel loss function for the prediction of RIASEC profiles, which captures the interrelationship between RIASEC dimensions. Finally, we profiled a set of job posts using \textsc{JPLink} and showed that our model managed to identify a type of imperfection, existed in the current profiling system.

There might be other imperfections in the current system (e.g.,  imperfections in mapping between different occupation taxonomies), which could be identified via a deep analysis of our predictions with the knowledge of domain experts. %Such an extensive study might open ways to make our model more sophisticated. 
Also, our model can be used to understand individuals' behaviors using the job posts that they prefer to apply. For instance, if individuals tend to apply for jobs with similar interest profiles, our model might be used to infer individuals' interest profile, which is currently collected via surveys. Likewise, we believe that our effort will open the door for many promising research directions. 
\vspace{-2mm}

\bibliographystyle{splncs04}
\bibliography{paper433-main}

\begin{thebibliography}{10}
\providecommand{\url}[1]{\texttt{#1}}
\providecommand{\urlprefix}{URL }
\providecommand{\doi}[1]{https://doi.org/#1}

\bibitem{burges2005learning}
Burges, C., Shaked, T., Renshaw, E., Lazier, A., Deeds, M., Hamilton, N.,
  Hullender, G.N.: Learning to rank using gradient descent. In: Proc. of ICML
  (2005)

\bibitem{cao2007learning}
Cao, Z., Qin, T., Liu, T.Y., Tsai, M.F., Li, H.: Learning to rank: from
  pairwise approach to listwise approach. In: Proc. of ICML (2007)

\bibitem{gati1998using}
Gati, I.: Using career-related aspects to elicit preferences and characterize
  occupations for a better person--environment fit. Journal of Vocational
  Behavior  (1998)

\bibitem{grover2016node2vec}
Grover, A., Leskovec, J.: node2vec: Scalable feature learning for networks. In:
  Proc. of KDD (2016)

\bibitem{holland1997making}
Holland, J.L.: Making vocational choices: A theory of vocational personalities
  and work environments. Psychological Assessment Resources (1997)

\bibitem{mikolov2013distributed}
Mikolov, T., Sutskever, I., Chen, K., Corrado, G.S., Dean, J.: Distributed
  representations of words and phrases and their compositionality. In: Proc. of
  NIPS (2013)

\bibitem{mislove2010you}
Mislove, A., Viswanath, B., Gummadi, K.P., Druschel, P.: You are who you know:
  inferring user profiles in online social networks. In: Proc. of WSDM (2010)

\bibitem{perozzi2014deepwalk}
Perozzi, B., Al-Rfou, R., Skiena, S.: Deepwalk: Online learning of social
  representations. In: Proc. of KDD (2014)

\bibitem{prediger_1981}
Prediger, D.J.: Mapping occupations and interests: A graphic aid for vocational
  guidance and research. Vocational Guidance Quarterly  (1981)

\bibitem{prediger1982dimensions}
Prediger, D.J.: Dimensions underlying holland's hexagon: Missing link between
  interests and occupations? Journal of Vocational Behavior  (1982)

\bibitem{prediger1992locating}
Prediger, D.J., Vansickle, T.R.: Locating occupations on holland's hexagon:
  Beyond riasec. Journal of Vocational Behavior  (1992)

\bibitem{rounds1999development}
Rounds, J., Smith, T., Hubert, L., Lewis, P., Rivkin, D.: Development of
  occupational interest profiles for {O*NET}. Raleigh, NC: National Center for
  {O*NET} Development  (1999)

\bibitem{tang2015line}
Tang, J., Qu, M., Wang, M., Zhang, M., Yan, J., Mei, Q.: Line: Large-scale
  information network embedding. In: Proc. of WWW (2015)

\bibitem{yamada2016joint}
Yamada, I., Shindo, H., Takeda, H., Takefuji, Y.: Joint learning of the
  embedding of words and entities for named entity disambiguation. Proc. of
  SIGNLL  (2016)

\bibitem{zhu2017document}
Zhu, Y., Javed, F., Ozturk, O.: Document embedding strategies for job title
  classification. In: Proc. of FLAIRS (2017)

\end{thebibliography}

\end{document}